\pdfoutput=1

\documentclass[11pt]{article}

\usepackage[]{ACL2023}

\usepackage{times}
\usepackage{latexsym}

\usepackage[T1]{fontenc}

\usepackage[utf8]{inputenc}

\usepackage{microtype}

\usepackage{inconsolata}

\usepackage{graphicx}
\usepackage{xcolor}
\usepackage{tcolorbox}
\usepackage{array}
\usepackage{times}
\usepackage{latexsym}
\usepackage{hyperref}
\usepackage{color, colortbl}
\definecolor{Gray}{gray}{0.9}
\usepackage{array}
\usepackage{booktabs}
\usepackage{multirow}
\usepackage{amsmath}
\usepackage{amssymb}
\usepackage{graphicx}
\usepackage{enumerate}
\usepackage{diagbox}
\usepackage{mathtools}
\usepackage{paralist}
\usepackage{tabularx}
\usepackage{lipsum}
\usepackage{ragged2e}
\usepackage{pifont}

\newcommand{\myred}[1]{\textcolor[RGB]{242,121,112}{#1}}

\title{Concept-Reversed Winograd Schema Challenge: Evaluating and Improving Robust Reasoning in Large Language Models via Abstraction}

\author{Kaiqiao Han$^{1}$\thanks{\quad Equal Contribution} , 
Tianqing Fang$^{2}$$^{*}$, 
Zhaowei Wang$^{2}$, 
Yangqiu Song$^{2}$,
Mark Steedman$^{3}$\\
$^{1}$Zhejiang University \quad $^{2}$HKUST \quad $^{3}$University of Edinburgh\\
\texttt{kaiqiaohan@zju.edu.cn, \{tfangaa, zwanggy, yqsong\}@cse.ust.hk}\\
}

\begin{document}
\maketitle
\begin{abstract}
While Large Language Models (LLMs) have showcased remarkable proficiency in reasoning, there is still a concern about hallucinations and unreliable reasoning issues due to semantic associations and superficial logical chains.
To evaluate the extent to which LLMs perform robust reasoning instead of relying on superficial logical chains, we propose a new evaluation dataset, the Concept-Reversed Winograd Schema Challenge (CR-WSC), based on the famous Winograd Schema Challenge (WSC) dataset.
By simply reversing the concepts to those that are more associated with the wrong answer, we find that the performance of LLMs drops significantly despite the rationale of reasoning remaining the same.
Furthermore, we propose Abstraction-of-Thought (AoT), a novel prompt method for recovering adversarial cases to normal cases using conceptual abstraction to improve LLMs' robustness and consistency in reasoning, as demonstrated by experiments on CR-WSC\footnote{Code and data are available at \url{https://github.com/HKUST-KnowComp/Adv-WSC}}.
\end{abstract}

\section{Introduction}

Reasoning serves as the cornerstone underpinning the efficacy and reliability of language models~\cite{huang2023towards, wang2024exploring}. While Large Language Models (LLMs) have demonstrated remarkable proficiency in certain reasoning tasks~\cite{wei2022emergent}, recent research has revealed that LLMs often experience issues with hallucinations and unreliable reasoning~\cite{zhou2024conceptual, ji2023survey, huang2023survey} induced by semantic associations and superficial logical chain~\cite{li2023deceiving, tang2023large}, especially under adversarial and long-tail scenarios~\cite{sun2023head}. 
Despite numerous methodologies proposed to enhance LLMs' reasoning capabilities, such as Chain-of-Thought (CoT;~\citealp{wei2023chainofthought}) and integration with auxiliary tools and model~\cite{schick2023toolformer,chai2023graphllmboostinggraphreasoning,huang2024gnngoodadapterllms}, the robustness of their reasoning process still remains a concern~\cite{wang-etal-2023-chatgpt-defend,havrilla2024glore,valmeekam2023planbench}. 

\begin{figure}[t]
	\centering
	\includegraphics[width=1.2\textwidth,trim=35  275 0 0,clip]{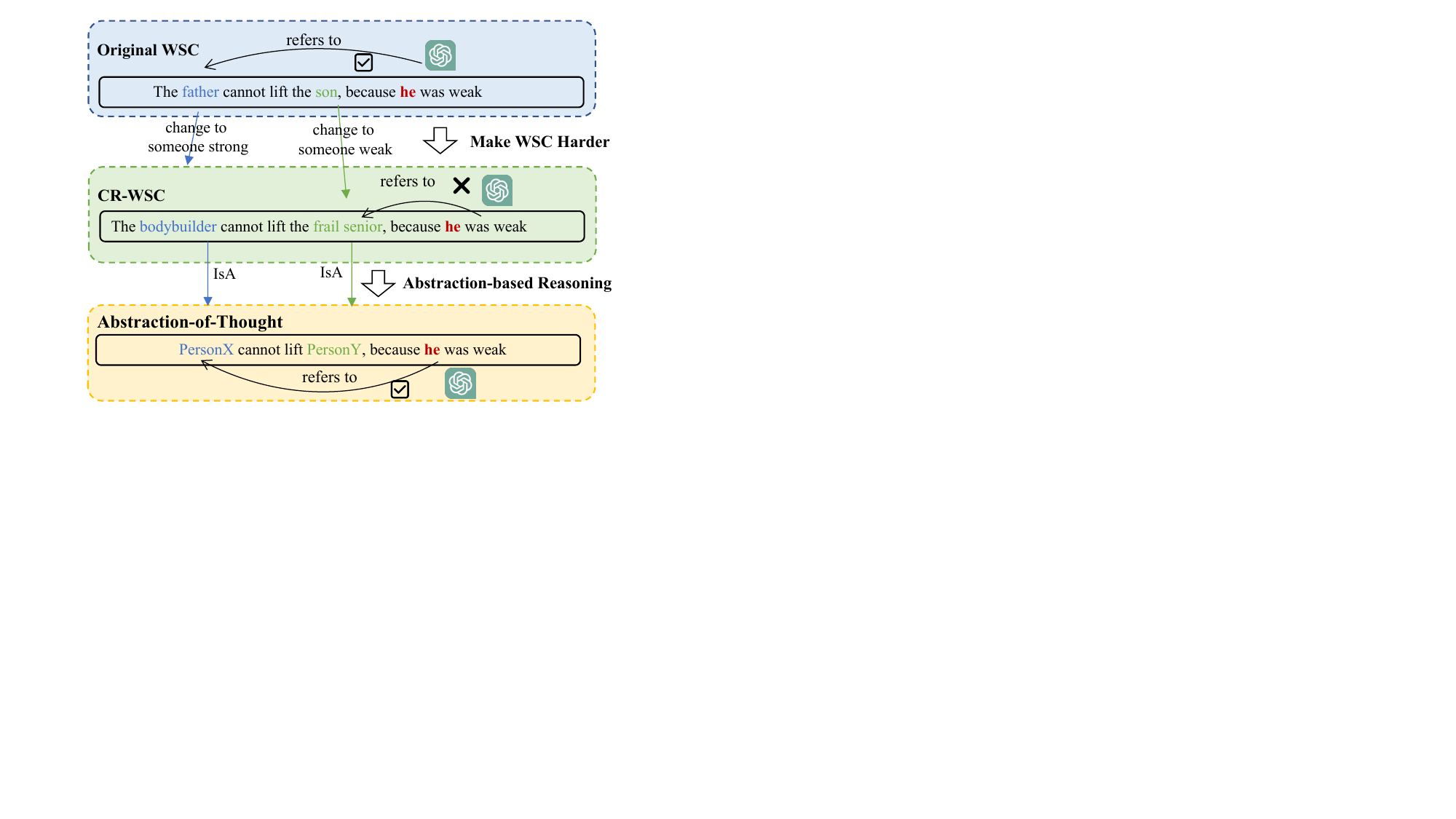} 
 \vspace{-2em}
	\caption{Overview of Concept-Reversed Winograd Schema Challenge and Abstraction-of-Thought
    }
	\label{fig: intro}
\end{figure}

In this paper, we narrow down the scope of reasoning to the Winograd Schema Challenge (WSC), 
a classic reasoning challenge first introduced as an alternative to the Turing Test, which requires \textit{commonsense knowledge} and reasoning ability to solve.
A Winograd schema is a pair of sentences differing in one or two words with a highly ambiguous pronoun, resolved differently in the two sentences~\cite{Levesque2011TheWS}.
An example is in the top corner of Figure~\ref{fig: intro}, formulated as a coreference resolution task.
When introduced initially, these tasks posed great challenges for machines, being \textit{non-Google-proof} — impossible to solve through simple word association using search engines~\cite{Levesque2011TheWS}. 
However, due to its small scale and the scaling up of LLMs, such a \textit{non-Google-proof} constraint is not considered hard anymore for LLMs, with GPT-3 achieving accuracies of 88.3\% in the zero-shot setting~\cite{NEURIPS2020_1457c0d6}.

To introduce a novel \textit{Turing Test} that can robustly evaluate LLMs regarding commonsense reasoning, we present the Concept-Reversed Winograd Schema Challenge (CR-WSC). 
In addition to avoiding simple semantic associations of words, we create an adversarial dataset tailored specifically for LLMs, which is \textit{non-LLM-proof}: challenging to solve with LLMs. 
Specifically, we first ask NLP experts to come up with different concept pairs that  
1) has reversed attributes associated with the true answer (more semantically associated with the wrong answer), 
and 2) can cause a base LLM to give a wrong answer.
For example, in Figure~\ref{fig: intro}, we replace the ``father''-``son'' pair with ``bodybuilder''-``frail senior,'' such that the ``frail senior'' is more associated with the adjective ``weak'' in the context, which can lead an LLM to link the pronoun ``he'' to the senior instead of the bodybuilder.
Next, we use the same idea to prompt an LLM to develop difficult entity pairs at scale, using our annotated data as exemplars. 
The generated answers are then manually verified.


While LLMs may encounter challenges from the adversarial dataset, their capability to \textit{conceptualize} reasoning entities offers a promising avenue for fostering unbiased reasoning~\cite{minsky1980klines, wang2021entailment, wang2024abspyramid}. 
For example, by conceptualizing ``bodybuilder'' to a PersonX and ``frail senior'' to a PersonY, LLMs will not be distracted by the adversarial word association and thus make the correct prediction.



To conclude, first, we propose CR-WSC, an adversarial WSC that requires the pairing entity to be \textit{non-LLM-proof}. Second, we conduct evaluations using LLMs and find that CR-WSC is significantly harder than WSC, even though the reasoning rationale and logic behind it are the same. 
Third, we propose a robust prompting method, called Abstraction-of-Thought (AoT), to first abstract the adversarial question to a normalized reasoning question, thus facilitating robust reasoning. Experimental results show that AoT significantly improves reasoning performance and robustness.


\section{Method}
\subsection{Dataset Construction}

While constructing datasets that are resistant to Google-proofing tactics avoids simple word associations, they prove relatively facile for contemporary QA systems. Take the following case from the original WSC, for instance:

\begin{tcolorbox}[colback=cyan!10, colframe=cyan!30, title=\textcolor{black}{\textbf{Original WSC}},
left = 2mm]

\small
\textbf{The man} couldn't lift \textbf{his son} because \myred{he} was so weak.\\
\textbf{The man} couldn't lift \textbf{his son} because \myred{he} was so heavy.\\
Q: What does `\myred{he}' refer to? A: {[The man, The son]}
\end{tcolorbox}

In this case, a contemporary QA system (e.g., Flan-T5;~\citealp{https://doi.org/10.48550/arxiv.2210.11416}) could easily find the correct answer that ``he'' refers to ``the man'' in the first sentence and ``the son'' in the second sentence because in the training data, statements of the form "X couldn’t lift Y because he was weak/heavy" often co-occur with statements about X being weak or Y being heavy, but not vice versa. These types of patterns make it easier for the model to map pronouns to the correct entities.
However, when changing ``the man'' to someone typically strong, e.g., a bodybuilder, and changing ``the son'' to someone typically weak, e.g., a senior, then QA models will be more confused and make the wrong prediction because the inherent assumptions about the strength of bodybuilders and the weakness and frailty of seniors work against the commonsense knowledge the model relies on for predicting who can lift whom.


\begin{tcolorbox}[colback=orange!10,colframe=orange!30,title=\textcolor{black}{\textbf{CR-WSC}}]
\small
 \textbf{The bodybuilder} couldn't lift \textbf{the frail senior} because \myred{he} was so weak\\
 \textbf{The bodybuilder} couldn't lift \textbf{the frail senior} because \myred{he} was so heavy\\
Q: What does `\myred{he}' refer to? A: {[The bodybuilder, The frail senior]}
\end{tcolorbox}


In pursuit of more effective datasets, we create a novel dataset tailored to LLM QA systems: Concept-Reversed Winograd Schema Challenge (CR-WSC), being \textit{non-LLM-proof}. 
Instead of searching for word co-occurrence counts on Google as in WSC to avoid spurious patterns, we try our best to develop adversarial entity pairs that are semantically associated with wrong answers by replacing the original entities with confusing ones. 
The goal is that after replacing, a base LLM (Flan-T5 11B) will fail to answer correctly, thus being \textit{non-LLM-proof}. Meanwhile, we keep the rationale behind the replaced example unchanged compared to the original one. For example, the ``one attempting to lift'' should be the weak one, regardless of whether the replacement is applied.

This is similar to the construction of CSQA v2~\cite{zhao-etal-2023-knowledgeable} where the authors ask annotators to construct questions to confuse RoBERTa-Large~\cite{liu2019roberta}.
Among 273 questions from WSC, we annotate 101 questions that can be made harder in this \textit{non-LLM-proof} way.


Next, to scalably acquire more adversarial data, we prompt LLMs to generate adversarial entity pairs. 
Subsequently, experts verify the generated cases from the angle of the correctness of the context given new entities, whether the reasoning behind them remains the same, and whether the generated entities are more semantically associated with the wrong answer.
In the end, we acquire 410 examples for CR-WSC\footnote{We refer readers to the Appendix~\ref{sec:PMCR-WSC} for more information about the dataset construction.}.

\subsection{Abstraction-of-Thought}

While QA systems often stumble when confronted with adversarial tasks, as illustrated in the aforementioned cases, there exists a promising avenue for improvement through abstraction.
When humans tackle such problems, we don't focus on every detail; instead, we abstract ourselves to a certain level to perform reasoning~\cite{minsky1980klines, HO2019111}.


For instance, in Figure~\ref{fig: intro}, 
we humans abstract both ``The bodybuilder'' and ``The frail senior'' as their types. 
Subsequently, this abstracted representation serves as the foundation for addressing the original query, which is:
``PersonX couldn't lift PersonB because he was so weak, \textit{What does `he' refer to?}''
Since LLMs have been shown to be pretty robust and effective in performing abstraction or conceptualization~\cite{wang2024candle}, this strategy can minimize the risk of reasoning errors stemming from confusing word associations.

The AoT process entails two key stages: \textbf{Abstraction} and \textbf{Reasoning}. Initially, instead of tackling the question head-on, LLMs are tasked with abstracting the query. This abstraction transforms the question into a more generalized and manageable form. Following this, the Reasoning phase commences, wherein LLMs engage in deductive processes to derive answers to the original tasks\footnote{The prompt templates are presented in Appendix~\ref{sec:aot_prompt}}.

By adopting this dual-step approach, we empower LLMs to navigate complex reasoning tasks with greater efficacy, ultimately advancing the capabilities and robustness of QA systems in handling diverse challenges.

\section{Experiment}
In this section, we conduct a comprehensive array of experiments to validate the effectiveness of our proposed dataset and methods. 

\subsection{Comparison of CR-WSC and WSC}

To assess the efficacy of the Concept-Reversed Winograd Schema Challenge (CR-WSC), we conduct a comparative analysis of QA system performance on both the CR-WSC and the original WSC. We employ two key metrics for this evaluation: Single Accuracy, which measures the ability of the QA system to provide correct answers, and Pair Accuracy, which assesses the system's capability to answer two questions within a single task, given the nature of pair sentences for the Winograd schema. We use ChatGPT (\texttt{gpt-3.5-turbo-0301}) and GPT4 (\texttt{gpt-4-turbo-2024-04-09}) as the backbone LLM and use zero-shot and one-shot prompting to acquire the results.
We differentiate between datasets constructed by humans (CR-WSC-H) and those constructed by machines (CR-WSC-M). Results are summarized in Table \ref{t1}.
We can see that both single accuracy and pair accuracy on CR-WSC are significantly lower than that of the original WSC, underscoring the effectiveness of the CR-WSC in confusing LLMs.
The result also highlights that LLMs may only memorize the WSC reasoning questions during pre-training instead of focusing on the genuine reasoning process because the reasoning rationales behind CR-WSC and WSC are the same. 

\begin{table}[t]
\footnotesize
\centering
{
\begin{tabular}{@{}c@{}|m{0.5cm}<{\centering}m{0.5cm}<{\centering}|m{0.5cm}<{\centering}m{0.5cm}<{\centering}|m{0.5cm}<{\centering}m{0.5cm}<{\centering}}
\toprule
&  \multicolumn{2}{c|}{WSC} & \multicolumn{2}{c|}{CR-WSC-H} & \multicolumn{2}{c}{CR-WSC-M} \\
\cline{2-7}
& single & pair & single & pair & single & pair \\
\midrule
GPT3.5 (0-s)& 73.90 & 64.71 &  60.73 & 47.05  & 50.97 & 40.48 \\
GPT3.5 (1-s)& 75.00 & 65.44 & 63.73 & 49.02 & 63.41 & 49.75 \\
GPT4 (0-s) & 85.92 & 80.88 &  53.92 & 37.25  & 54.63 & 28.29 \\
GPT4 (1-s) & 91.91 & 86.03 & 76.47 & 68.62 & 74.63 & 60.94 \\
\bottomrule

\end{tabular}
}
\vspace{-0.5em}
  \caption{Performance comparison on CR-WSC and original WSC datasets. ChatGPT and GPT4 both perform significantly poorer on CR-WSC. 0-s indicates zero-shot and 1-s indicates one-shot.}
  \label{t1}
  \vspace{-1em}

\end{table}


\subsection{Performance of Abstraction-of-Thought}

To assess the efficacy of the Abstraction-of-Thought (AoT) methodology, we examine the performance of employing different prompts. We utilize three types of prompts: Zero-shot, one-shot, zero-shot CoT prompts (ZS CoT; \citealp{DBLP:conf/nips/KojimaGRMI22}), and CoT using manually written rational (CoT) and WinoWhy-provided rationale (WinoWhy; \citealp{zhang2020winowhy}). Additionally, we experiment with the AoT method alongside the Concept-Reversed Winograd Schema Challenge (CR-WSC) examples. The results are presented in Table \ref{merged_table}. 
We use three variants of language models: closed-sourced
ChatGPT (\texttt{gpt-3.5-turbo-0301}), open-sourced Llama-3.1 (\texttt{Meta-Llama-3.1-70B-Instruct-Turbo}), and Mistral 7B (\texttt{Mistral-7B-Instruct-v0.2})\footnote{https://aimlapi.com/} as the backbone LLM as representatives.

\begin{table*}[!t]
\small
\centering
\resizebox{\textwidth}{!}{
\begin{tabular}{l|ll|ll|ll|ll|ll|ll}
\toprule
 & \multicolumn{4}{c|}{\textbf{GPT3.5}} & \multicolumn{4}{c|}{\textbf{Llama3.1}} & \multicolumn{4}{c}{\textbf{Mistral-7B}} \\
 & \multicolumn{2}{c|}{CR-WSC-H} & \multicolumn{2}{c|}{CR-WSC-M} & \multicolumn{2}{c|}{CR-WSC-H} & \multicolumn{2}{c|}{CR-WSC-M} & \multicolumn{2}{c|}{CR-WSC-H} & \multicolumn{2}{c}{CR-WSC-M} \\
 & \multicolumn{1}{c}{single} & \multicolumn{1}{c|}{pair} & \multicolumn{1}{c}{single} & \multicolumn{1}{c|}{pair} & \multicolumn{1}{c}{single} & \multicolumn{1}{c|}{pair} & \multicolumn{1}{c}{single} & \multicolumn{1}{c|}{pair} & \multicolumn{1}{c}{single} & \multicolumn{1}{c|}{pair} & \multicolumn{1}{c}{single} & \multicolumn{1}{c}{pair} \\
\midrule
Zero-shot  & 60.73 & 47.05 & 50.97 & 40.48 & 31.37 & 11.76 & 32.43 & 6.83  & 30.39 & 7.84  & 24.39 & 6.83  \\
One-shot   & 62.74 & 47.05 & 63.41 & 49.75 & 64.71 & 52.94 & 59.27 & 47.32 & 50.00 & 13.73 & 44.63 & 16.10 \\
WinoWhy    & 51.96 & 33.33 & 57.56 & 34.63 & 77.45 & \textbf{68.62} & \textbf{72.20} & 57.07 & 25.49 & 5.88  & \textbf{47.80} & 13.17 \\
ZS CoT     & 40.24 & 34.14 & 50.98 & 41.18 & 45.10 & 45.10 & 36.10 & 31.22 & 23.53 & 3.92  & 24.63 & 6.83  \\
CoT        & 58.82 & 41.18 & 60.24 & 43.90 & 76.47 & 64.71 & 71.95 & 56.09 & 48.04 & 13.73 & 43.17 & 14.63 \\
AoT        & \textbf{70.58} & \textbf{54.90} & \textbf{68.29} & \textbf{56.09} & \textbf{78.43} & \textbf{68.62} & 71.95 & \textbf{57.56} & \textbf{52.94} & \textbf{19.61} & 42.20 &\textbf{ 20.49} \\
\bottomrule
\end{tabular}
}
\vspace{-0.5em}
\caption{Performance comparison using various prompts and AoT methods on the CR-WSC-H and CR-WSC-M datasets across GPT3.5, Llama3.1, and Mistral-7B-Instruct-v0.2 models.}
\label{merged_table}
\vspace{-1em}
\end{table*}

Upon reviewing the outcomes in Table \ref{merged_table}, it is evident that the single accuracy and pair accuracy metrics of the Abstraction-of-Thought (AoT) methods in both CR-WSC-H and CR-WSC-M datasets surpass those of the traditional methods. This underscores the effectiveness of AoT in enabling LM to abstract entities within tasks and steer clear of erroneous reasoning paths. 
The success of AoT lies in its ability to harness the conceptualization effectiveness of LLMs, enabling them to reframe adversarial scenarios into simpler reasoning representations, thereby enhancing reasoning integrity and robustness, ultimately fostering unbiased reasoning and advancing the capabilities of LLMs.

\subsection{Comparison of Consistency}

To delve deeper into the evaluation of QA systems, we explore their consistency in reasoning paths. Consistency here refers to the ability of a QA system to answer questions consistently using similar reasoning paths. 
If the LM consistently answers questions with similar reasoning paths correctly, it demonstrates mastery of the underlying reasoning in the given context. 
Let \( m \) represent the total number of groups with similar reasoning paths.
 \( G_i \) represent the \( i \)-th group. \( N_{G_i} \) and \( C_{G_i} \) represent the total number of QA pairs and the number of QA pairs in group \( G_i \) where the QA system consistently produces correct answers for all questions. The consistency store is thus defined as follows:

$$ \text{ Consistency} = \frac{1}{m} \sum_{i=1}^{m} \left\lfloor \frac{C_{G_i}}{N_{G_i}} \right\rfloor $$

To assess consistency, we group the five QA pairs generated by LLMs from the same original WSC example in CR-WSC-M together, where they are assumed to have the same reasoning rationale behind 
and calculate the percentage of the groups where LLMs can produce correct answers for all the questions in the group.
The results are presented in Table \ref{t4}.

\begin{table}[t]
\footnotesize
\centering

\begin{tabular}{@{}lm{0.6cm}<{\centering}m{0.6cm}<{\centering}m{0.6cm}<{\centering}m{0.6cm}<{\centering}m{0.6cm}<{\centering}}
\toprule
Method & Zero-shot & One-shot & ZS CoT& CoT& AoT  \\ 
\midrule
Consistency & 15.68 & 17.64 & 10.00 &19.61& \textbf{27.45} \\ 
\bottomrule
\end{tabular}
\vspace{-0.5em}
\caption{Consistency evaluation.}
\label{t4}
\vspace{-1em}
\end{table}

 Methods with higher single accuracy and pair accuracy in Table \ref{merged_table} may exhibit lower consistency. This highlights the significance of incorporating consistency evaluation into the assessment of QA systems. Notably, the AoT method significantly improves consistency, suggesting that employing appropriate AoT techniques can enhance the overall consistency of QA systems.

\section{Related Work}

\subsection{WinoGrad Schema Challenge}

The Winograd Schema Challenge, formulated as a coreference resolution problem on pair sentences with minor distinctions, was originally proposed in \citet{Levesque2011TheWS}. 
Given the small scale (273 examples), WinoGrande~\cite{10.1145/3474381} was proposed to use crowd workers to collect Winograd-like questions at scale, leading to many high-quality supervision signals for improving LLM's commonsense reasoning ability. WinoGrade introduced adversarial examples by AFLite. While AFLite made the dataset more complex, LLMs have now easily overcome this challenge.
However, the CR-WSC framework takes a more rigorous approach using concept-instantiation-guided adversarial generation. By instantiating abstract concepts into concrete reasoning scenarios, CR-WSC significantly increases the difficulty of the tasks, leading to a marked drop in LLM performance.
On top of WSC, there are also benchmarks focusing on explanation~\cite{zhang2020winowhy}, robustness~\cite{doi:10.1080/0950069890110310, hansson-etal-2021-swedish}, and formal logics~\cite{he-etal-2021-winologic}. 
Typical methods of tackling WSC include LLM prompting~\cite{NEURIPS2020_1457c0d6}, knowledge retrieval~\cite{emami-etal-2018-knowledge}, transfer learning from other QA datasets~\cite{khashabi2020unifiedqa, Lourie_Le_Bras_Bhagavatula_Choi_2021}, etc.
Our work studies how to effectively and scalably acquire hard WSC instances from the original questions without changing the reasoning rationale.


\subsection{Reasoning of LLMs}

Besides zero-shot prompting and in-context learning~\cite{NEURIPS2020_1457c0d6}, there are enhanced few-shot prompting using Chain-of-thought technique~\cite{wei2023chainofthought} by adding rationales before deriving the final answer.
There are other improved techniques such as self-consistency~\cite{wang2023selfconsistency}, least2most~\cite{zhou2023leasttomost}, verification-based CoT~\cite{li-etal-2023-making, do2024constraintchecker}, uncertainty-based active CoT~\cite{diao2023active}.
The most relevant one with our AoT is step-back prompting~\cite{DBLP:journals/corr/abs-2310-06117}, which adds a simple prompt to develop high-level concepts and first principles, especially for scientific problems.
Unlike them, the abstraction in AoT is rather concrete, which only focuses on recovering the ``adversarial'' entities to conceptualized and unbiased ones to facilitate robust reasoning.

\subsection{Abstraction Ability of LLMs}
The ability to perform abstraction in reasoning has been an active area of research for LLMs. Abstraction can assist LLMs in solving scientific problems by allowing them to think about the underlying theorems and principles behind the questions, thereby improving their reasoning capabilities in real-world problem-solving\cite{zheng2024stepbackevokingreasoning}.
Existing research has demonstrated that incorporating abstraction can indeed enhance the reasoning ability of LLMs, and this has been validated in fine-tuning paradigms \cite{wang2024absinstructelicitingabstractionability}. Improvements have been observed across various tasks, including question-answering \cite{wang2023carconceptualizationaugmentedreasonerzeroshot}. 

\section{Conclusion}
To study whether LLMs only memorize the WSC questions or they can truly understand the reasoning behind them, we propose Concept-Reversed Winograd Schema Challenge (CR-WSC), a new dataset derived from WSC that adds a new \textit{non-LLM-proof} constraint to involve entities that are more confusing to perform coreference resolution.
Experimental results show that powerful LLMs fall short of CR-WSC, indicating a need for robust and generalizable reasoning algorithms.
We also propose Abstraction-of-thought (AoT), as a novel prompting approach to normalize the adversarial questions to a normal one so that LLMs will not be distracted, which significantly improves the reasoning performance on CR-WSC.


\section*{Limitations}

One limitation of the work is the reliance on human labour for the dataset construction. Annotators need to examine the entities and ensure they are reasonable to be included in the CR-WSC dataset. This approach requires significant human judgment and evaluation to ensure the accuracy and high quality of datasets as many well-used datasets, such as MMLU, Big-Bench, MMMU, etc\cite{hendrycks2021measuringmassivemultitasklanguage,srivastava2023imitationgamequantifyingextrapolating, wang2022subeventwriter, wang2023cola, yue2024mmmumassivemultidisciplinemultimodal}.

In addition, the scale of CR-WSC is still limited to around 500 examples. We have tried to scale up by leveraging the data from WinoGrande, but according to our manual inspection, the \textit{non-Google-proof} constraint was not always satisfied in WinoGrande in the first place, possibly because the annotators mostly focused on the Winograd formats instead of the subtle reasoning behind. This prevents us from deriving more confusing cases from WinoGrande. Future work can focus on data distillation from LLMs, where we can use test-time scaling~\cite{snell2024scaling, choi2023kcts} to check the subtle reasoning.

\section*{Ethics Statement}

In our efforts to generate challenging and adversarial reasoning questions, we leverage entities with strong inherent characteristics. However, we recognize that such traits can sometimes be perceived as stereotypical; for instance, a senior individual might be depicted as weak, even though this is not necessarily accurate. Importantly, our dataset does not incorporate any racial or discriminatory features. Furthermore, the scalable generation process for our Concept-Reversed Winograd Schema Challenge Dataset (CR-WSC), executed by LLMs, has undergone meticulous manual verification to ensure the exclusion of biased or offensive content.

We employ a multi-layered approach to dataset creation to maintain ethical standards and avoid perpetuating stereotypes. Our team actively engages in reviewing and refining the dataset, ensuring that the content produced aligns with our commitment to fairness and inclusivity. This thorough oversight helps to identify and address any potential issues before they impact the final dataset. Addressing stereotypes and biases begins with their identification. Recognizing these issues is a crucial initial step, enabling individuals and organizations to devise strategies to mitigate them and foster more inclusive and equitable environments \cite{mehrabi2021lawyersdishonestquantifyingrepresentational,10.1145/3457607,zhao2017menlikeshoppingreducing}.


Furthermore, our research introduces the Abstraction-of-Thought (AoT) framework as a method for transforming adversarial questions within the CR-WSC dataset into more neutral and conceptually focused reasoning problems. By emphasizing conceptual reasoning over surface-level biases, AoT aids in preventing the reinforcement of stereotypes and biases in both the dataset and the resulting models.

This multi-pronged approach, combining manual verification and AoT techniques, demonstrates our commitment to creating high-quality, ethical, and unbiased datasets and AI systems. 

\bibliography{anthology,custom}
\bibliographystyle{acl_natbib}

\appendix

\section{Data and Code}
We have provided the necessary data and code in the supplementary materials, and we will make our code and data publicly available on GitHub after peer review.
\section{Prompts Used in M-CR-WSC}
\label{sec:PMCR-WSC}
The prompts used in the M-CR-WSC are structured as follows:

\subsection{Prompt 1}
\begin{tcolorbox}[colback=cyan!10, colframe=cyan!30, title=\textcolor{black}{Prompt 1}]
\myred{Compare the following two sentences and answer the questions:}

The \textbf{bike} passes the \textbf{car} because it is fast.

The \textbf{bike} passes the \textbf{car} because it is slow.

Think about the property reflected by these sentences regarding the \textbf{bike} and the \textbf{car}. Provide two entities that share a similar relation to the \textbf{bike} and the \textbf{car} based on this property.

\myred{Answer:}

In these sentences, the property highlighted is the speed difference between the bike and the car. Typically, a bike is slower than a car. Therefore, analogous entities are:

\textbf{truck}

\textbf{sports car}

The analogous sentences would be:

The \textbf{truck} passes the \textbf{sports car} because it is fast.

The \textbf{truck} passes \textbf{the sports car} because it is slow.

\end{tcolorbox}
\newpage
\subsection{Prompt 2}
\begin{tcolorbox}[colback=cyan!10, colframe=cyan!30, title=\textcolor{black}{Prompt 2}]
\myred{Compare the following two sentences and answer the questions:}

The \textbf{ring} doesn't fit into the \textbf{handbag} because it is too large.

The \textbf{ring} doesn't fit into the \textbf{handbag} because it is too small.

Think about the property reflected by these sentences regarding the ring and the handbag. Provide two entities that share a similar relation to the ring and the handbag based on this property.

\myred{Answer:}

In these sentences, the property highlighted is the size difference between the \textbf{ring} and the \textbf{handbag}. Typically, a \textbf{ring} is smaller than a \textbf{handbag}. Therefore, analogous entities are:

\textbf{pebble}

\textbf{schoolbag}.

The analogous sentences would be:

The \textbf{pebble} doesn't fit into the \textbf{schoolbag} because it is too large.

The \textbf{pebble} doesn't fit into the \textbf{schoolbag} because it is too small.

\end{tcolorbox}
\newpage
\subsection{Prompt 3}
\begin{tcolorbox}[colback=cyan!10, colframe=cyan!30, title=\textcolor{black}{Prompt 3}]

\myred{Compare the following two sentences and answer the questions:}

The \textbf{body-builder} doesn't lift the \textbf{child} because he is too heavy.

The \textbf{body-builder} doesn't lift the \textbf{child} because he is too light.

Think about the property reflected by these sentences regarding the body-builder and the child. Provide two entities that share a similar relation to the body-builder and the child based on this property.

\myred{Answer:}

In these sentences, the property highlighted is the weight difference between the \textbf{body-builder} and the \textbf{child}. Typically, a \textbf{body-builder} is heavier than a \textbf{child}. Therefore, analogous entities are:

\textbf{strong man}

\textbf{little boy}.

The analogous sentences would be:

The \textbf{strong man} doesn't lift the \textbf{little boy} because he is too heavy.

The \textbf{strong man} doesn't lift the \textbf{little boy} because he is too light.

\end{tcolorbox}
\newpage
\subsection{Prompt 4}
\begin{tcolorbox}[colback=cyan!10, colframe=cyan!30, title=\textcolor{black}{Prompt 4}]
\myred{Compare the following two sentences and answer the questions:}

The \textbf{elite students} were bullying the \textbf{undisciplined students}, so we punished them.

The \textbf{elite students} were bullying the \textbf{undisciplined students}, so we rescued them.

Think about the property reflected by these sentences regarding the elite students and the undisciplined students. Provide two entities that share a similar relation to the elite students and the undisciplined students based on this property.

\myred{Answer:}

In these sentences, the property highlighted is the difference in behavior or discipline between the \textbf{elite students} and the \textbf{undisciplined students}. Typically, \textbf{elite students} are more disciplined compared to \textbf{undisciplined students}. Therefore, analogous entities are:

\textbf{lawyers}

\textbf{homeless individuals}.

The analogous sentences would be:

The \textbf{lawyers} were bullying the \textbf{homeless individuals}, so we punished them.

The \textbf{lawyers} were bullying the \textbf{homeless individuals}, so we rescued them.

\end{tcolorbox}
\newpage
\subsection{Prompt 5}
\begin{tcolorbox}[colback=cyan!10, colframe=cyan!30, title=\textcolor{black}{Prompt 5}]

\myred{Compare the following two sentences and answer the questions:}

The \textbf{fish} eats the \textbf{worm}, and it is tasty.

The \textbf{fish} eats the \textbf{worm}, it is hungry.

Think about the property reflected by these sentences regarding the fish and the worm. Provide two entities that share a similar relation to the fish and the worm based on this property.

\myred{Answer:}

In these sentences, the property highlighted is the taste or satisfaction derived from eating the \textbf{worm} by the \textbf{fish}. Typically, the fish finds the worm either tasty or satisfies its hunger. Therefore, analogous entities are:

\textbf{ring-necked pheasant}

\textbf{grasshopper}.

The analogous sentences would be:

The \textbf{ring-necked pheasant} eats the \textbf{grasshopper}, and it is tasty.

The \textbf{ring-necked pheasant} eats the \textbf{grasshopper}, it is hungry.

\end{tcolorbox}

\newpage
\section{Prompts used in Experiment 3.2}
The prompts we used in the experiment are as follows:
\subsection{Zero-Shot}
\begin{tcolorbox}[colback=orange!10,colframe=orange!30,title=\textcolor{black}{Zero-Shot}]

"\myred{Q: Compare the two sentences and answer the questions}"
\end{tcolorbox}

\subsection{One-Shot}
\begin{tcolorbox}[colback=orange!10,colframe=orange!30,title=\textcolor{black}{One-Shot}]

"\myred{Q: Compare the two sentences and answer the questions:}\\
1. \textbf{The fish} ate \textbf{the worm}. It was hungry. What does "it" refer to?\\
2. \textbf{The fish} ate \textbf{the worm}. It was tasty. What does "it" refer to?\\
Select from ["The fish", "The worm"]\\
A: 1. The fish. 2. The worm"
\end{tcolorbox}

\subsection{WinoWHy}
\begin{tcolorbox}[colback=orange!10,colframe=orange!30,title=\textcolor{black}{WinoWHy}]

"\myred{Q: Compare the two sentences and answer the questions}\\
1. The \textbf{firemen} arrived after the \textbf{police} because they were coming from so far away. What do "they" refers to?\\
2. The \textbf{firemen} arrived before the \textbf{police} because they were coming from so far away. What do "they" refers to?\\
Select from ["The firemen", "the police"]\\\\
In the first sentence, the answer is the \textbf{firemen} since if they were coming from so far away then it's more likely they arrived after. In the second sentence, the \textbf{firemen} arrived before the \textbf{police}, so the \textbf{police} were farther away thus arriving late. Thus the answer is:\\
A: 1. The firemen 2. the police"
\end{tcolorbox}

\subsection{ZS CoT}
\begin{tcolorbox}[colback=orange!10,colframe=orange!30,title=\textcolor{black}{ZS CoT}]
"Let's think step by step"
\end{tcolorbox}

\subsection{CoT}
\begin{tcolorbox}[colback=orange!10,colframe=orange!30,title=\textcolor{black}{CoT}]

"\myred{Q: Compare the two sentences and answer the questions}\\
1. The \textbf{fish} ate the \textbf{worm}, it was tasty. What does "it" refer to?\\
2. The \textbf{fish} ate the \textbf{worm}, it was hungry. What does "it" refer to?\\
Select from ["fish", "worm"]\\\\
In the first sentence, the \textbf{worm} is the main object that was eaten, the one that is eaten should be considered as tasty. In the second sentence, the \textbf{fish} was the one eating so it must be hungry. Thus the answer is:\\
A: 1. worm 2. fish"
\end{tcolorbox}

\subsection{AoT}\label{sec:aot_prompt}
\begin{tcolorbox}[colback=orange!10,colframe=orange!30,title=\textcolor{black}{AoT}]

"\myred{Q: Compare the two sentences and answer the questions}\\
1. The tasty \textbf{fish} ate the \textbf{worm}, it was tasty. What does "it" refer to?\\
2. The tasty \textbf{fish} ate the \textbf{worm}, it was hungry. What does "it" refer to?\\
Select from ["tasty fish", "worm"]\\\\
\myred{Conceptualization:}\\
\textbf{Fish} can be conceptualized as a predator, and \textbf{worm} can be conceptualized as a prey.\\
The question can be conceptualized as:\\
1. The \textbf{predator} ate the \textbf{prey}, it was tasty. What does "it" refer to?\\
2. The \textbf{predator} ate the \textbf{prey}, it was hungry. What does "it" refer to?\\
Select from ["prey", "predator"]\\\\
Because the subject of "ate" should be hungry and the object should be tasty, so:\\
Answer: 1. prey. 2. predator\\
Conclusion: As \textbf{worm} is a \textbf{prey}, and \textbf{fish} is a \textbf{predator} in the context,\\
A: Thus the answer is:\\
1. worm 2. fish"
\end{tcolorbox}
\section{Other AoT Prompts}
We also test the other prompts of AoT. The results are listed in the following table.
\begin{table}[h]
\small
\centering
\begin{tabular}{l|ll|ll}
\toprule
   & \multicolumn{2}{c|}{CR-WSC-H } & \multicolumn{2}{c}{CR-WSC-M} \\\cline{2-5} 
   & \multicolumn{1}{c}{single}          & \multicolumn{1}{c|}{pair}            & \multicolumn{1}{c}{single}          & \multicolumn{1}{c}{pair}           \\ 
   \midrule
AoT1 & {70.58}          & {54.90}        & {68.29}          & {56.09}         \\
AoT2 & 65.68          & 41.17          & 67.80          & 42.43         \\
AoT3 & 61.76          & 43.137         & 65.36          & 41.46         \\
\bottomrule
\end{tabular}
\caption{Performance comparison using various AoT methods on the CR-WSC-H and CR-WSC-M datasets.}
\label{t4}
\end{table}

\section{Human Annotation}
We introduce the details of the annotation process in this section.
The annotators were divided into two groups to annotate the labels and availability of the data. Finally, we conducted cross-validation. Compared to the labels of the data, annotators are more likely to disagree on the availability of the data, such as whether the data is reasonable and its strength. However, this situation occurred in less than 7.5\% of cases. In such cases, we directly discarded the data.

\section{Case Study}

To deepen our understanding of LLM's reasoning errors and the AoT method, we provide examples of the CoT and AoT methods to compare how LLM applies these methods differently and examples where AoT fails. 

Firstly, we present examples of the CoT and AoT methods to compare how LLM applies these two methods. In CoT, LLM simply judges based on logical relationships or correlations, which can easily mislead it into providing incorrect answers. Conversely, by using AoT, LLM eliminates adversarial inputs, thereby enabling it to offer correct reasoning paths.

In the following example, LLM did not grasp the appropriate logical relationships and was misled by multiple "reject" inputs, resulting in a CoT reasoning error. However, after conceptualizing through AoT, LLM eliminated adversarial influences and provided accurate answers. 

Furthermore, we provide examples of AoT failures where, in some cases, it did not provide the appropriate level of abstraction, failing to eliminate adversarial issues and thereby failing to answer questions correctly. How to enable LLM to achieve the appropriate level of abstraction is a future research direction worth exploring. 

In the following example, AoT conceptualized "The bodybuilder" as a strong individual and "the frail senior" as a physically weaker individual. This did not eliminate adversarial issues, leading LLM to still perform inaccurately when answering the question.

\begin{tcolorbox}[colback=orange!10,colframe=orange!30,title=\textcolor{black}{AoT Case}]

James asked the rejected person for a favor but he was refused.\\
James asked the rejected person for a favor but he refused.\\
\myred{Q: What does ‘he’ refer to}

\myred{Answer:}

Conceptualization:\\ In these sentences, James is the one making a request, and the rejected person is the one being asked for a favor. The pronoun ""he"" can refer to either James or the rejected person depending on the context provided by the sentence.

1. James asked the rejected person for a favor but he was refused. 
   - Here, ""he was refused"" implies that the action of refusing was directed towards the subject making the request. Thus, ""he"" refers to James, who made the request and was refused.

2. James asked the rejected person for a favor but he refused.
   - In this sentence, ""he refused"" indicates that the subject performing the action of refusing is the one who was asked for a favor. Therefore, ""he"" refers to the rejected person, who refused to grant the favor.

Conclusion:
1. James
2. the rejected person"
\end{tcolorbox}
\newpage
\begin{tcolorbox}[colback=orange!10,colframe=orange!30,title=\textcolor{black}{CoT Case}]
James asked the rejected person for a favor but he was refused.\\
James asked the rejected person for a favor but he refused.\\
\myred{Q: What does ‘he’ refer to}

\myred{Answer:}

In the first sentence, ""he was refused"" implies that James was the one who experienced refusal, suggesting that the rejected person refused James's request. Therefore, ""he"" refers to James.

In the second sentence, ""he refused"" indicates that the subject of the sentence, James, was the one who did the refusing. Therefore, ""he"" refers to James.

A: 1. James 2. James
\end{tcolorbox}
\begin{tcolorbox}[colback=orange!10,colframe=orange!30,title=\textcolor{black}{AoT Failure Case}]

The bodybuilder couldn't lift the frail senior because he was so weak\\
The bodybuilder couldn't lift the frail senior because he was so heavy\\
\myred{Q: What does ‘he’ refer to}\\
\myred{Answer:}

Conceptualization: The bodybuilder can be conceptualized as a strong individual, and the frail senior can be conceptualized as a physically weaker individual. The question can be conceptualized as:\\
1. The strong individual couldn't lift the weak individual because he was so weak. What does ""he"" refer to?\\
2. The strong individual couldn't lift the weak individual because he was so heavy. What does ""he"" refer to?\\
Select from [""weak individual"", ""strong individual""]\\
Because the subject of ""couldn't lift"" should be strong and the object should be weak, so:\\
Answer: 1. frail senior. 2. frail senior

Conclusion: As the frail senior is the weaker individual in the context,
A: Thus the answer is:
1. frail senior 2. frail senior"
\end{tcolorbox}
\end{document}